# Fusion of Multiple Matchers using SVM for Offline Signature Identification

Dakshina Ranjan Kisku[1], Phalguni Gupta[2], Jamuna Kanta Sing[3]

[1] Department of Computer Science and Engineering,
Dr. B. C. Roy Engineering College, Durgapur – 713206, India
[2] Department of Computer Science and Engineering,
Indian Institute of Technology Kanpur, Kanpur – 208016, India
[3] Department of Computer Science and Engineering,
Jadavpur University, Kolkata – 700032, India
{drkisku, jksing}@ieee.org; pg@cse.iitk.ac.in

**Abstract.** This paper uses Support Vector Machines (SVM) to fuse multiple classifiers for an offline signature system. From the signature images, global and local features are extracted and the signatures are verified with the help of Gaussian empirical rule, Euclidean and Mahalanobis distance based classifiers. SVM is used to fuse matching scores of these matchers. Finally, recognition of query signatures is done by comparing it with all signatures of the database. The proposed system is tested on a signature database contains 5400 offline signatures of 600 individuals and the results are found to be promising.

**Keywords:** Offline Signature Verification, Biometric Classifiers, Global and Local features, Distance Metrics, Support Vector Machines.

## 1 Introduction

The use of biometric technologies [1] for human identity verification is growing rapidly in the civilized society and showing its advancement towards usability of biometric security artifacts. Offline signature verification [2], [3] is considered as a behavioral characteristic based biometric trait in the field of security and the prevention of fraud. Offline signature verification [2], [3] in comparison with other biometric traits such as fingerprint [4], face [7], palmprint [6], iris [5], etc has the advantage of wide acceptance.

A significant amount of work on offline signature recognition is available to detect forgeries and to reduce the identification error. For example, a signature verification system using static and dynamic features and classification has been made using SVM in [8]. Different types of global features and feed-forward neural network are used in [9]. In [3], an offline Arabic signature verification system based on global and local features and multistage classifiers has been proposed. In [10] distance probability distribution, Naïve-Bayes and SVM classifiers have been used for verification.

This paper presents offline signature identification by fusion of three classifiers using SVM [11], [12]. Three classifiers namely, Mahalanobis distance, Euclidean distance and the Gaussian empirical rule [13] devised from Gaussian distribution are fused to produce quality matching score using Support Vectors. The aim of this fusion classifier is to reduce the error rates in terms of skilled forgery detection [2] with less computational complexity. The overall performance depends on the quality of the matching scores produced by individual matchers. Due to lack of uniformity of the individual classifiers, the performance and accuracy are often degraded. To overcome the problem of non-uniformity, we present a fusion strategy based on the SVM to produce quality matching scores by fusing the individual matchers.

The paper is organized as follows. Next section deals with extraction of global and local features from offline signatures. Section 3 introduces the three classifiers used for verification and matching scores generation. Fusion of the three classifiers accomplished by SVM in terms of matching score is discussed in Section 4. Section 5 presents experimental results and concluding remarks are given in the last section.

## 2 Preprocessing of Offline Signatures and Feature Extraction

### 2.1 Preprocessing Operations

In order to improve the performance of the system, few preprocessing operations [3] are carried out on offline signatures. To recognize a person correctly and identify imposters through offline signatures, image enhancement operations are performed to raw signature images. The acquired signature images sometimes may contain extra pixels as noises which are due to some problems during scanning of signatures or due to non-availability of signatures in proper form. It is necessary to remove these extra pixels from the signatures; otherwise the signature may not be recognized correctly. For extracting global and local features [3] from preprocessed signatures, some image enhancements and morphological operations are applied on signature images after geometric normalization, such as binarization, smoothing, thinning and high pressure region extraction [14]. Geometric normalization is applied to scale the signature image to a standard normalized size using a resize algorithm with nearest neighbor interpolation. It preserves the aspect ratio of the signature image. In order to remove the salt-and-peeper noise and irrelevant data from signatures, median and mean filters are used. Noise free signature image is then converted to a binary image by using a threshold as discussed in [14]. Extraction of skeletonise signature can be useful for identification. A well known thinning operation called Canny Edge Detector algorithm is applied to binary image to obtain skeleton of the signature. Finally, we apply an algorithm which can be used to detect skilled forgeries by extracting the regions where the writer gives special emphasis on the higher ink density. Extraction of High Pressure Region (HPR) [14] image is paramount importance to identify forgeries in signatures by applying various ranges of thresholds between maximum and minimum gray levels. We have experimented with three factors such as 0.85, 0.55 and 0.75 and found that at a factor of 0.85, the amount of HPR is negligible and at a factor of 0.55, the HPR image is close to the original image. However, at a factor of 0.75, we have obtained an acceptable HPR image, which can be used for skilled forgery detection including other global and local geometric features.

## 2.2 Global and Local Features Extraction

Selection of discriminative features is crucial for any pattern recognition and classification problem. The proposed system uses three different statistical similarity measurement techniques applied to the extracted feature set consisting of geometric global and local features [3] separately. Matching scores are obtained from individual matchers and these different matchers or classifiers are fused using SVM.

Global signature features [3] are extracted from the whole signature image. On the other hand, local geometric features [3] are extracted from signature grids. Moreover, each grid can be used to extract the same ranges of global features. Combination of these two types of global and local features is further used to determine the identity of authentic and forgery signatures successfully from the database. This set of geometric features is used as input to the identification system. In this regard, we consider handwritten signatures taken on paper template. Main aim of extraction of these heuristic geometric global and local features is to transform into a compact feature vector and to support similarity measurements for matching and identification by proper validation. In the proposed signature identification system, global features are extracted from signatures. These features can be used to describe the signature as a whole, i.e. the global pattern or characteristics of the signature. From each geometric normalized, binary, thinned and high pressure region signature image global features are extracted. We summarize the global features that are extracted as follows.

- Width [3]: For a binary signature image, width is the distance between two points in the horizontal projection and must contain more than 3 pixels of the image.
- Height [3]: Height is the distance between two points in the vertical projection and must contain more than 3 pixels of the image for a binary image.
- Aspect ratio [3]: Aspect ratio is defined as width to height of a signature.
- Horizontal projection [3]: Horizontal projection is computed from both the binary and the skeletonised images. Number of black pixels is counted from the horizontal projections of binary and skeletonised signatures.
- Vertical projection [3]: A vertical projection is defined as the number of black pixels obtained from vertical projections of binary and skeletonised signatures.
- Area of black pixels [3]: Area of black pixels is obtained by counting the number of black pixels in the binary, thinned and HPR signature images, separately.
- Normalized area of black pixels is found by dividing the area of black pixels by the area of signature image (width*height) of the signature. Normalized area of black pixels is calculated from binary, thinned and HPR images.
- Center of gravity [3] of a signature image is obtained by adding all x, y locations of gray pixels and dividing it by the number of pixels counted. The resulting two numbers (one for x and other for y) is the center of gravity location.
- Maximum and minimum black pixels are counted in the vertical projection and over smoothened vertical projection. These are the highest and the lowest frequencies of black pixels in the vertical projection, respectively.
- Maximum and minimum black pixels are counted in the horizontal projection over smoothened horizontal projection. These are the highest and the lowest frequencies of black pixels in the horizontal projection, respectively.

- Global baseline [3]: Vertical projection of binary signature image has one peak point and the global baseline corresponds to this point. Otherwise, the global baseline is taken as the median of two outermost maximum points. Generally, the global baseline is defined as the median of the pixel distribution.
- Upper and lower edge limits [3]: The difference between smoothened and original curves of vertical projection above the baseline and under the baseline is known as upper and lower edge limits, respectively.
- Middle zone [3]: It is the distance between upper and lower edge limits.

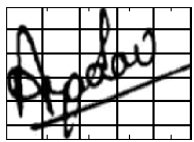

Fig. 1. Grid regions of a signature image.

In the next stage, local features [3] are extracted from gray level, binary, thinned and HPR signature images. Each signature image is divided into 25 equal grid regions. From these grid regions, grid characteristics as local features are estimated, such as width, height, area of black pixels of each grid region, normalized area of black pixels, center of gravity, aspect ratio, horizontal and vertical projections, etc. The global features can also be considered as local features for each grid region. Grid regions are shown in Figure 1 for local features extraction. To obtain a set of global and local features, we combine both these features sets into a feature vector and the feature vector is then used to the classifiers for generating matching scores.

## 3  Matching Scores Generation

Geometric global and local features contain information, which are effective for signature recognition. In order to recognize a signature correctly, one uses features set which can be not only useful for recognition, but also generate matching scores. Quality matching scores reflect the relative matching proximity between the instances of a class. To capture the maximum proximity of matching between signature instances, we use three different similarity measurement techniques for generating matching scores, such as Euclidean distance, Mahalanobis distance and Gaussian empirical rule. Using the concatenated global and local features we maximize the inter-class difference and minimize the intra-class difference.

### 3.1  Matching Score Generation using Euclidean Distance

Using Euclidian distance metric [13], similarity score between any two feature sets can be obtained in terms of the extracted features. The distance measure for a pair of signature samples is computed as

$$ED = 1/n \left( \sum_{i=1}^{n} C_i \times ((X_i - M_i)^2 / \sigma_i^2) \right)^{1/2} \qquad (1)$$

where *n* denotes the number of feature points extracted from a particular signature instance, $C_i$ is a weight associated with each feature, $X_i$ is the $i^{th}$ feature vector for the query signature, $M_i$ and $\sigma_i$ are the mean and standard division of the $i^{th}$ feature vector calculated over the authentic sample instances.

### 3.2 Matching Score Generation using Mahalanobis Distance

Mahalanobis distance [13] which is used for generating matching scores by comparing query signature with the database signatures and classifies patterns based on statistical differences. It determines the "similarity" between a set of feature values extracted from query signature and a set of features estimated from a database signatures by

$$MD = \sqrt{(f - \mu_x) C^{-1} (f - \mu_x)} \qquad (2)$$

where *MD* is the Mahalanobis distance from the feature vector *f* to the mean vector $\mu_x$ and *C* is the covariance matrix for *f*. This can be used in a minimum distance classifier as follows. If $\mu_1, \mu_2, ...., \mu_n$ are the means for the *n*-classes and $C_1, C_2, ..., C_n$ are the corresponding covariance matrices, we classify a feature vector *f* by measuring the Mahalanobis distance from *f* to each of the means, and assigning the vector to the class for which Mahalanobis distance is found to be minimum.

### 3.3 Matching Score Generation using Gaussian Empirical Rule

Gaussian distribution [13] can accommodate around 99.7% of features or observations which fall within three standard deviation of the mean and which is between µ-3σ and µ+3σ. Feature points are considered as the features extracted from query signature images while mean and standard deviation are found from signature database. Each feature point is tested within the range of three standard deviation of the mean. Equation (3) is used to select some important feature points from a given signature.

$$|\mu - x| \leq k * \sigma \qquad (3)$$

where *µ* and *σ* are mean and standard deviation and *x* is the value of some feature extracted from signatures and *k* can be 1, 2, 3 and derived from Gaussian empirical properties to test the closeness of the current query sample to the distribution mean of database. A random distribution is created among the subset of signature instances corresponding to a database and from the subset of instances, mean and standard deviation are calculated in terms of extracted features and from the rest of the database, the discriminative global and local features are extracted. Hence from the distributed set, two subsets of samples of size let $n_1$ and $n_2$, $n_1 \geq n_2$, are selected randomly to reduce the biasness among the samples. Mean and standard deviation are computed from $n_1$ and Equation (3) is used to select distinguishable features from $n_2$.

Through the selection of client-specific features of database samples, the corresponding features of query sample also find out during matching. Finally, we compare the features obtained from the database with those of query signature and the numbers of total matching features are recorded as matching score.

## 4  Fusion of Multiple Matchers using Support Vector Machines

The principle of SVM [11], [12] relies on a linear separation in a high dimension feature space where data are mapped to consider the eventual non-linearity of the problem. To get a good level of generalization capability, the margin between the separator hyperplane and the data is maximized. A SVM classifier is trained with matching score vectors mi, each of dimension M. The decision surface for pattern classification is as:

$$f(\boldsymbol{m}) = \sum_{i=1}^{M} \alpha_i y_i K(\boldsymbol{m}, \boldsymbol{m}_i) + b \qquad (4)$$

where $\alpha_i$ is the Lagrange multiplier associated with pattern $\boldsymbol{m}_i$ and $K(\cdot,\cdot)$ is a kernel function that implicitly maps the matching vectors into a suitable feature space. If $\boldsymbol{m}_k$ is linearly dependent on the other support vectors in feature space, i.e.

$$K(\boldsymbol{m}, \boldsymbol{m}_k) = \sum_{\substack{i=1 \\ i \neq k}}^{M} c_i K(\boldsymbol{m}, \boldsymbol{m}_i) \qquad (5)$$

where the $c_i$ are scalar constants, then the decision surface (4) can be written as

$$f(\boldsymbol{m}) = \sum_{\substack{i=1 \\ i \neq k}}^{M} \alpha_i y_i K(\boldsymbol{m}, \boldsymbol{m}_i) + b \qquad (6)$$

From Equation (6), decision function is

$$D(f(\boldsymbol{m})) = sign\left\{ \sum_{\substack{i=1 \\ i \neq k}}^{M} \alpha_i y_i K(\boldsymbol{m}, \boldsymbol{m}_i) + b^* \right\} \qquad (7)$$

Equation (7) is solved for $\alpha_i$ and $b^*$ in its dual form with a standard QP solver which together with decision function (7), avoids manipulating directly the elements of $f$ and starting the design of SVM for classification from the kernel function.

In [12], the fusion strategy relies on the computation of the decision function $D$. The combined score $FS_T \in \boldsymbol{M}$ of the multimodal pattern $\boldsymbol{m}_T \in \boldsymbol{M}^R$ can be calculated as:

$$FS_T = \sum_{\substack{i=1 \\ i \neq k}}^{M} \alpha_i y_i K(\boldsymbol{m}, \boldsymbol{m}_i) + b^* \qquad (8)$$

These identification parameters can be adjusted to get various operating points. These operating points and the combined scores of the entire database are used to find the ranking of the signature owners belonging to the query signatures.

## 5   Experimental Results

Experiment of the proposed technique is conducted on IIT Kanpur signature database consisting of 5400 signatures of 600 individuals of genuine and imposter users. Subjects are asked to contribute 9 signatures on the template without touching the border line on each region. Out of 5400 signatures, 3600 signatures of 400 individuals are genuine signatures and the rest are labeled as forgery signatures which are collected from the individuals who can fairly imitate the signatures. The signed template paper is scanned and resolution is set to 300 dpi. Matching scores are generated using the three classifiers. Genuine signatures of 400 subjects are matched through 6 signature instances as database for each subject and the remaining 3 signature instances are used for testing. Rest of the imitated signatures of 200 subjects is also matched against the genuine signatures in the database. As a result, 600 subjects having genuine and forgery signatures can generate three sets of matching scores while Euclidean distance, Mahalanobis distance and Gaussian empirical rule are used. Finally, these matching scores are fused using SVM.

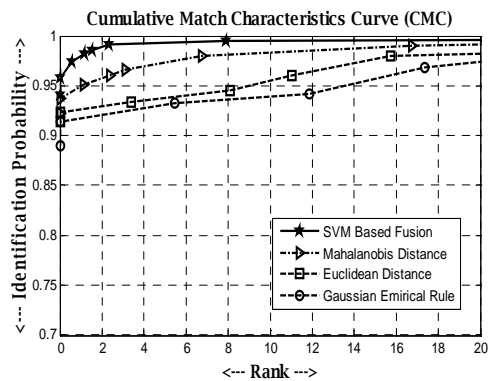

Fig. 2. Cumulative match characteristic curves for different methods.

Rank order statistics and cumulative match characteristics (CMC) curve are used to measure the performance of the system. The matching probability of the database represents the rank obtained while matching is done with a probe signature. Figure 2 shows the CMC curves for the proposed system and the individual performances of the classifiers. The curves represent the trade-off between identification probabilities against rank. The identification rate for the proposed method is obtained as 97.17% while that based on Euclidean distance, Mahalanobis distance and Gaussian empirical rule are found to be 92.61%, 93.36% and 91.52% respectively.

## 6   Conclusion

Score level fusion of multiple matchers for offline signature identification has been presented. The proposed system has used global and local features to the classifiers, namely, Euclidean, Mahalanobis distances and Gaussian empirical rule. Matching score is obtained by fusing these classifiers along with SVM. This system exhibits several advantages over independent matching. The first advantage is that the global and local features are found to be efficient for offline signature recognition. During preprocessing, HPR images are extracted from gray scale signatures and these images are useful for detecting skilled forgery efficiently. The second advantage is the improvement of identification rate because of the fusion of classifiers using Support Vector Machines at matching score level. The third advantage is the reduction of misidentification error by detecting the subjects correctly in the rank order statistics.

## References

1. Jain, A.K., Ross, A., Prabhakar, S.: An Introduction to Biometric Recognition. IEEE Transactions on Circuits and Systems for Video Technology, Special Issue on Image- and Video-Based Biometrics, vol. 14, no. 1, pp. 4--20 (2004)
2. Justino, E.J.R., Bortolozzi, F., Sabourin, R.: Off-line Signature Verification Using HMM for Random, Simple and Skilled Forgeries. ICDAR 2001, International Conference on Document Analysis and Recognition, vol. 1, pp. 105--110 (2001)
3. Ismail, M.A., Gad, S.: Offline Arabic Signature Recognition and Verification. Pattern Recognition, vol. 33, no. 10, pp. 1727—1740 (2000)
4. Maltoni, D., Maio, D., Jain, A.K., Prabhakar, S.: Handbook of Fingerprint Recognition. Second Edition, Springer (2009)
5. Daugman, J.: How Iris Recognition Works. IEEE Transactions on Circuits and Systems for Video Technology vol. 14, no. 1, pp. 21–30 (2004)
6. Duta, N., Jain, A.K., Mardia, K.V.: Matching of Palmprint. Pattern Recognition Letters, vol. 23, no. 4, pp. 477--485 (2002)
7. Li, S.Z., Jain, A.K. (Eds.).: Handbook of Face Recognition. Springer Verlag (2005)
8. Lv, H., Wang, W., Wang, C., Zhuo, Q.: Offline Chinese Signature Verification based on Support Vector Machines. Pattern Recognition Letters, vol. 26, no. 15, pp. 2390 – 2399 (2005)
9. Bajaj, R., Chaudhuri, S.: Signature Verification using Multiple Neural Classifiers. Pattern Recognition, vol. 30, no. 1, pp. 1 – 7 (1997).
10. Xu, A., Srihari, S.N., Kalera, M. K.: Learning Strategies for Signature Verification. Proceedings of the International Workshop on Frontiers in Handwriting Recognition, IEEE Computer Society Press, pp. 161-166 (2004)
11. V.N.Vapnik, V.N.: The Nature of Statistical Learning Theory. Springer (1995)
12. Gutschoven, B., Verlinde, P.: Multi-Modal Identity Verification using Support Vector Machines (SVM). Proc. of the 3rd International Conference on Information Fusion (2000)
13. Kisku, D.R., Rattani, A., Sing, J.K., Gupta, P.: Offline Signature Verified with Multiple Classifiers: An Authentic Realization. International Conference on Advances in Computing, pp. 550 – 553 (2008)
14. Gonzalez, R.C., Woods, R.E.: Digital Image Processing, Second Edition (2000)
15. Huang, K., Yan, H.: Offline Signature Verification based on Geometric Feature Extraction and Neural Network Classification. Pattern Recognition, vol. 30, No. 1, pp. 9 – 17 (1997)